\documentclass[11pt,letterpaper,twocolumn]{article}

\usepackage[paperwidth=8.5in,paperheight=11.0in,
  left=1.0in,right=1.0in,top=1.0in,bottom=1.0in,
  heightrounded]{geometry}
  
\usepackage{url}
\usepackage{setspace}
\usepackage[numbers]{natbib}
\begin{document}

\onecolumn
\begin{center} 
\large \textbf{The Debate Over Understanding in AI’s Large Language Models}

\vspace*{.2in}

\normalsize
Melanie Mitchell and David C. Krakauer \\ 
\vspace*{.1in}
Santa Fe Institute, 1399 Hyde Park Road, Santa Fe, NM 87501 \\
mm@santafe.edu, krakauer@santafe.edu

\vspace*{.1in}

\end{center}

\begin{abstract}
We survey a current, heated debate in the AI research community on whether large pre-trained language models can be said to understand language---and the physical and social situations language encodes---in any humanlike sense.  We describe arguments that have been made for and against such understanding, and key questions for the broader sciences of intelligence that have arisen in light of these arguments.  We contend that an extended science of intelligence can be developed that will provide insight into distinct modes of understanding, their strengths and limitations, and the challenge of integrating diverse forms of cognition. 
\end{abstract}

\onehalfspacing
What does it mean to understand something?  
This question has long engaged philosophers, cognitive scientists, and educators, nearly always with reference to humans and other animals. However, with the recent rise of large-scale AI systems---especially so-called large language models---a heated debate has arisen in the AI community on whether machines can now be said to understand natural language, and thus understand the physical and social situations that language can describe.  This debate is not just academic; the extent and manner in which machines understand our world has real stakes for how much we can trust them to drive cars, diagnose diseases, care for the elderly, educate children, and more generally act robustly and transparently in tasks that impact humans. Moreover, the current debate suggests a fascinating divergence in how to think about understanding in intelligent systems, in particular the contrast between mental models that rely on statistical correlations and those that rely on causal mechanisms. 

Until quite recently there was general agreement in the AI research community about machine understanding: while AI systems exhibit seemingly intelligent behavior in many specific tasks, they do not \textit{understand} the data they process in the way humans do. Facial recognition software does not understand that faces are parts of bodies, or the role of facial expressions in social interactions, or what it means to ``face'' an unpleasant situation, or any of the other uncountable ways in which humans conceptualize faces.  Similarly, speech-to-text and machine translation programs do not understand the language they process, and autonomous driving systems do not understand the meaning of the subtle eye contact or body language drivers and pedestrians use to avoid accidents. Indeed, the oft-noted \textit{brittleness} of these AI systems---their unpredictable errors and lack of robust generalization abilities---are key indicators of their lack of understanding \cite{mitchell2019artificial}.  However, over the last several years, a new kind of AI system has soared in popularity and influence in the research community, one that has changed the views of some people about the prospects of machines that understand language.  Variously called Large Language Models (LLMs), Large Pre-Trained Models, or Foundation Models \cite{bommasani2021opportunities}, these systems are deep neural networks with billions to trillions of parameters (weights) that are ``pre-trained'' on enormous natural-language corpora, including large swathes of the Web, online book collections, and other collections amounting to terabytes of data.  The task of these networks during training is to predict a hidden part of an input sentence---a method called ``self-supervised learning.''  The resulting network is a complex statistical model of how the words and phrases in its training data correlate.   Such models can be used to generate natural language, be fine-tuned for specific language tasks \cite{min2021recent}, or be further trained to better match ``user intent'' \cite{ouyang2022training}. LLMs such as OpenAI's well-known GPT-3 \cite{brown2020language} (and more recent ChatGPT \cite{schulman2022chatgpt}) and Google's PaLM \cite{chowdhery2022palm} can produce astonishingly humanlike text, conversation, and, in some cases, what seems like human reasoning abilities \cite{wei2022chain}, even though the models were not explicitly trained to reason.  How LLMs perform these feats remains mysterious for lay people and scientists alike. The inner workings of these networks are largely opaque; even the researchers building them have limited intuitions about systems of such scale.  The neuroscientist Terrence Sejnowski described the emergence of LLMs this way: ``A threshold was reached, as if a space alien suddenly appeared that could communicate with us in an eerily human way.  Only one thing is clear---LLMs are not human...Some aspects of their behavior appear to be intelligent, but if not human intelligence, what is the nature of their intelligence?'' \cite{sejnowski2022large}.

As impressive as they are, state-of-the-art LLMs remain susceptible to brittleness and unhumanlike errors.  However, the observation that such networks improve significantly as their number of parameters and size of training corpora are scaled up \cite{wei2022emergent} has led some in the field to claim that LLMs---perhaps in a multi-modal version---will lead to human-level intelligence and understanding, given sufficiently large networks and training datasets. A new AI mantra has emerged: ``Scale is all you need.'' \cite{defreitas2022tweet,dimakis2022tweet}. 

Such claims are emblematic of one side of the stark debate in the AI research community on how to view LLMs.  One faction argues that these networks truly understand language and can perform reasoning in a general way (though ``not yet'' at the level of humans).  For example, Google's LaMDA system, which was pre-trained on text and then fine-tuned on dialogue \cite{thoppilan2022lamda}, is sufficiently convincing as a conversationalist that it convinced one AI researcher that such systems ``in a very real sense understand a wide range of concepts'' \cite{aguera2021understand} and are even ``making strides towards consciousness'' \cite{aguera2022consciousness}.  Another machine language expert sees LLMs as a canary in the coal mine of general human-level AI: ``There is a sense of optimism that we are starting to see the emergence of knowledge-imbued systems that have a degree of general intelligence'' \cite{manning2022understanding}. Another group argues that LLMs ``likely capture important aspects of meaning, and moreover work in a way that approximates a compelling account of human cognition in which meaning arises from conceptual role'' \cite{piantasodi2022meaning}.  Those who reject such claims are criticized for promoting ``AI denialism'' \cite{aguera2022behave}.

Those on the other side of this debate argue that large pretrained models such as GPT-3 or LaMDA---however fluent their linguistic output---cannot possess understanding because they have no experience or mental models of the world; their training in predicting words in vast collections of text has taught them the \textit{form} of language but not the meaning \cite{bender2020climbing,bender2021dangers,marcus2022nonsense}.  A recent opinion piece put it this way: ``A system trained on language alone will never approximate human intelligence, even if trained from now until the heat death of the universe'' and ``it is clear that these systems are doomed to a shallow understanding that will never approximate the full-bodied thinking we see in humans'' \cite{browning2022limits}.  Another scholar argued that intelligence, agency, and by extension, understanding ``are the wrong categories'' for talking about these systems; instead LLMs are compressed repositories of human knowledge more akin to libraries or encyclopedias than to intelligent agents \cite{gopnikAI2022}.  For example, humans know what is meant by a “tickle” making us laugh, because we have bodies. An LLM could use the word "tickle", but it has obviously never had the sensation.  Understanding a tickle is to map a word to a sensation, not to another word. 

Those on the ``LLMs do not understand'' side of the debate argue that while the fluency of large language models is surprising, our surprise reflects our lack of intuition of what statistical correlations can produce at the scales of these models.  Anyone who attributes understanding or consciousness to LLMs is a victim of the Eliza effect \cite{hofstadter1995ineradicable}---named after the 1960s chatbot created by Joseph Weizenbaum that, simple as it was, still fooled people into believing it understood them \cite{weizenbaum1976computer}.  More generally, the Eliza effect refers to our human tendency to attribute understanding and agency to machines with even the faintest hint of humanlike language or behavior.

A 2022 survey given to active researchers in the natural-language-processing community shows the stark divisions in this debate.  One survey item asked if the respondent agreed with the following statement about whether LLMs could ever, in principle, understand language: ``Some generative model [i.e., language model] trained only on text, given enough data and computational resources, could understand natural language in some non-trivial sense.'' Of 480 people responding, essentially half (51\%) agreed, and the other half (49\%) disagreed \cite{michael2022nlp}.

Those who would grant understanding to current or near-future LLMs base their views on the performance of these models on several measures, including subjective judgement of the quality of the text generated by the model in response to prompts (though such judgements can be vulnerable to the Eliza effect), and more objective performance on benchmark datasets designed to assess language understanding and reasoning.  For example, two standard benchmarks for assessing LLMs are the General Language Understanding Evaluation (GLUE) \cite{wang2018glue}, and its successor (SuperGLUE) \cite{wang2019superglue}, which include large-scale datasets with tasks such as ``textual entailment'' (given two sentences, can the meaning of the second be inferred from the first?), ``words in context'' (does a given word have the same meaning in two different sentences?), and yes/no question answering, among others.  OpenAI's GPT-3, with 175 billion parameters, performed surprisingly well on these tasks \cite{brown2020language}, and Google's PaLM, with 540 billion parameters, performed even better \cite{chowdhery2022palm}, often equaling or surpassing humans on the same tasks.  

What do such results say about understanding in LLMs?  The very terms used by the researchers who named these benchmark assessments---``general language understanding,'' ``natural language inference,'' ``reading comprehension,'' ``commonsense reasoning,'' and so on---reveal an assumption that humanlike understanding is required to perform well on these tasks.  But do these tasks actually require such understanding?  Not necessarily.  As an example, consider one such benchmark, the Argument Reasoning Comprehension Task \cite{habernal2018argument}.  In each task example, a natural-language ``argument'' is given, along with two statements; the task is to determine which statement is consistent with the argument.  Here is a sample item from the dataset:

\begin{quote}
\textbf{Argument:} Felons should be allowed to vote.  A person who stole a car at 17 should not be barred from being a full citizen for life.

\textbf{Statement A:} Grand theft auto is a felony.

\textbf{Statement B:} Grand theft auto is not a felony.
\end{quote}

An LLM called BERT \cite{devlin2019bert} obtained near-human performance on this benchmark \cite{niven2019probing}.  It might be concluded that BERT understands natural-language arguments as humans do.  However, one research group discovered that the presence of certain words in the statements (e.g., ``not'') can help predict the correct answer.  When researchers altered the dataset to prevent these simple correlations, BERT's performance dropped to essentially random guessing \cite{niven2019probing}.  This is a straightforward example of ``shortcut learning''---a commonly cited phenomenon in machine learning in which a learning system relies on spurious correlations in the data, rather than humanlike understanding, in order to perform well on a particular benchmark \cite{geirhos2020shortcut,gururangan2018annotation,lapuschkin2019unmasking,mccoy2019right}. Typically such correlations are not apparent to humans performing the same tasks.  While shortcuts have been discovered in several standard benchmarks used to evaluate language understanding and other AI tasks, many other, as yet undetected, subtle shortcuts likely exist. Pre-trained language models at the scale of Google's LaMDA or PaLM models---with hundreds of billions of parameters, trained on text amounting to billions or trillions of words---have an unimaginable ability to encode such correlations.  Thus benchmarks or assessments that would be appropriate for measuring human understanding might not be appropriate for assessing such machines \cite{choudhury2022machine,gardner2021competency,linzen2020can}.  It is possible that, at the scale of these LLMs (or of their likely near-future successors), any such assessment will contain complex statistical correlations that enable near-perfect performance without humanlike understanding.  

While ``humanlike understanding'' does not have a rigorous definition, it does not seem to be based on the kind of massive statistical models that today's LLM's learn; instead it is based on \textit{concepts}---internal mental models of external categories, situations, and events, and of ones own internal state and ``self''.  In humans, understanding language (as well as nonlinguistic information) requires having the concepts that language (or other information) describes, beyond the statistical properties of linguistic symbols.  Indeed, much of the long history of research in cognitive science has been a quest to understand the nature of concepts, and how understanding arises from coherent, hierarchical sets of relations among concepts that include underlying causal knowledge \cite{baumberger2017understanding,kvanvig2018knowledge}. These models enable people to abstract their knowledge and experiences in order to make robust predictions, generalizations, and analogies, to reason compositionally and counterfactually, to actively intervene on the world in order to test hypotheses, and to explain one's understanding to others \cite{goldwater2015acquisition,gopnik2022causal,hofstadter2013surfaces,keil2006explanation,lake2017building,sloman2015causality,smolensky2022neurocompositional}.  Indeed, these are precisely the abilities lacking in current AI systems, including state-of-the-art LLMs, though ever-larger LLMs have exhibited limited sparks of these general abilities.  It has been argued that understanding of this kind may enable abilities not possible for purely statistical models \cite{de2004discussion,george2020captcha,lake2021word,pearl2018theoretical,strevens2013no}.   While LLMs exhibit extraordinary \textit{formal linguistic competence}---the ability to generate grammatically fluent, humanlike language---they still lack the conceptual understanding needed for humanlike \textit{functional} language abilities---the ability to robustly understand and use language in the real world \cite{mahowald2023dissociating}.   An interesting parallel can be made between this kind of functional understanding and the success of formal mathematical techniques applied in physical theories \cite{krakauer2020limits}. For example, a long-standing criticism of quantum mechanics is that it provides an effective means of calculation without providing conceptual understanding.

The detailed nature of human concepts has been the subject of active debate for many years.  Researchers disagree on the extent to which concepts are domain-specific and innate versus more general-purpose and learned \cite{carey1995origin,goodman2011learning,gopnik2011unified,mandler1992build,spelke2007core,wellman1992cognitive}, the degree to which concepts are grounded via embodied metaphors \cite{gibbs2017metaphor,lakoff1980metaphorical,murphy1996metaphoric}, are represented in the brain via dynamic, situation-based simulations \cite{barsalou2008grounded}, and the conditions under which concepts are underpinned by language \cite{de2014role,dove2020more,lupyan2016language}, by social learning \cite{akhtar2000social,waxman2009early,gelman2009learning} and by culture \cite{bender2017causal,morris2003culturally,norenzayan2000culture}. In spite of these ongoing debates, concepts, in the form of causal mental models as described above, have long been considered to be the units of understanding in human cognition.  Indeed, the trajectory of human understanding---both individual and collective---is the development of highly compressed, causally based models of the world, analogous to the progression from Ptolemy's epicycles to Kepler's elliptical orbits, and to Newton's concise and causal account of planetary motion in terms of gravity.  Humans, unlike machines, seem to have a strong innate drive for this form of understanding, both in science and in everyday life \cite{gopnik1994theory}. We might characterize this form of understanding as requiring little data, minimal or parsimonious models, clear causal dependencies, and strong mechanistic intuition.  
 
The key questions of the debate about understanding in LLMs are the following: (1) Is talking of understanding in such systems simply a category error, mistaking associations between language tokens for associations between tokens and physical, social, or mental experience?  In short, is it the case that these models are not, and will never be, the kind of things that can understand?  Or conversely, (2) do these systems (or will their near-term successors) actually, even in the absence of physical experience, create something like the rich concept-based mental models that are central to human understanding, and, if so, does scaling these models create ever better concepts?  Or (3) If these systems do not create such concepts, can their unimaginably large systems of statistical correlations produce abilities that are functionally equivalent to human understanding?  Or, indeed, that enable new forms of higher-order logic that humans are incapable of accessing?  And at this point will it still make sense to call such correlations ``spurious'' or the resulting solutions ``shortcuts?''  And would it make sense to see the systems' behavior not as ``competence without comprehension'' but as a new, nonhuman form of understanding? These questions are no longer in the realm of abstract philosophical discussions, but touch on very real concerns about the capabilities, robustness, safety, and ethics of AI systems that increasingly play roles in humans' everyday lives.

While adherents on both sides of the ``LLM understanding'' debate have strong intuitions supporting their views, the cognitive-science-based methods currently available for gaining insight into understanding are inadequate for answering such questions about LLMs.  Indeed, several researchers have applied psychological tests---originally designed to assess human understanding and reasoning mechanisms---to LLMs, finding that LLMs do, in some cases, exhibit humanlike responses on theory-of-mind tests \cite{aguera2021understand,trott2022large} and humanlike abilities and biases on reasoning assessments \cite{binz2022using,dasgupta2022language,laverghetta2022predicting}.  While such tests are thought to be reliable proxies for assessing more general abilities in humans,they may not be so for AI systems. As we described above, LLMs have an unimaginable capacity to learn correlations among tokens in their training data and inputs, and can use such correlations to solve problems for which humans, in contrast, seem to apply compressed concepts that reflect their real-world experiences.  When applying tests designed for humans to LLMs, interpreting the results can rely on assumptions about human cognition that may not be true at all for these models.  To make progress, scientists will need to develop new kinds of benchmarks and probing methods that can yield insight into the mechanisms of diverse types of intelligence and understanding, including the novel forms of "exotic, mind-like entities" \cite{shanahan2022talking} we have created, perhaps along the lines of some promising initial efforts \cite{li2021implicit,olsson2022context}.    

The debate over understanding in LLMs, as ever larger and seemingly more capable systems are developed, underscores the need for a extending our sciences of intelligence in order to make sense of broader conceptions of understanding, for both humans and machines.  As neuroscientist Terrence Sejnowski points out, ``The diverging opinions of experts on the intelligence of LLMs suggests that our old ideas based on natural intelligence are inadequate'' \cite{sejnowski2022large}. If LLMs and related models succeed by exploiting statistical correlations at a heretofore unthinkable scale, perhaps this could be considered a novel form of ``understanding'', one that enables extraordinary, superhuman predictive ability, such as in the case of the AlphaZero and AlphaFold systems from DeepMind \cite{jumper2021highly,silver2017mastering}, which respectively seem to bring an ``alien'' form of intuition to the domains of chess playing and protein-structure prediction \cite{jones2022impact,sadler2019game}. 

It could thus be argued that in recent years the field of AI has created machines with new modes of understanding, most likely new species in a larger zoo of related concepts, that will continue to be enriched as we make progress in our pursuit of the elusive nature of intelligence. And just as different species are better adapted to different environments, our intelligent systems will be better adapted to different problems.  Problems that require enormous quantities of historically encoded knowledge where performance is at a premium will continue to favor large-scale statistical models like LLMs, and those for which we have limited knowledge and strong causal mechanisms will favor human intelligence. The challenge for the future is to develop new scientific methods that can reveal the detailed mechanisms of understanding in distinct forms of  intelligence, discern their strengths and limitations, and learn how to integrate such truly diverse modes of cognition. 

\subsection*{Acknowledgments}
This material is based in part upon work supported by the National Science Foundation under Grant No. 2020103. Any opinions, findings, and conclusions or recommendations expressed in this material are those of the author and do not necessarily reflect the views of the National Science Foundation.

% Bibliography
\bibliography{ai}

\begin{thebibliography}{85}
\providecommand{\natexlab}[1]{#1}
\providecommand{\url}[1]{\texttt{#1}}
\expandafter\ifx\csname urlstyle\endcsname\relax
  \providecommand{\doi}[1]{doi: #1}\else
  \providecommand{\doi}{doi: \begingroup \urlstyle{rm}\Url}\fi

\bibitem[Aguera~y Arcas(2021)]{aguera2021understand}
B.~Aguera~y Arcas.
\newblock Do large language models understand us?, 2021.
\newblock Medium, December 16, \url{tinyurl.com/38t23n73}.

\bibitem[Aguera~y Arcas(2022{\natexlab{a}})]{aguera2022behave}
B.~Aguera~y Arcas.
\newblock Can machines learn how to behave?, 2022{\natexlab{a}}.
\newblock Medium, August 3, \url{tinyurl.com/mr4cb3dw}.

\bibitem[Aguera~y Arcas(2022{\natexlab{b}})]{aguera2022consciousness}
B.~Aguera~y Arcas.
\newblock Artificial neural networks are making strides towards consciousness,
  2022{\natexlab{b}}.
\newblock The Economist, June 13, \url{tinyurl.com/ymhk37uu}.

\bibitem[Akhtar and Tomasello(2000)]{akhtar2000social}
N.~Akhtar and M.~Tomasello.
\newblock The social nature of words and word learning.
\newblock In \emph{Becoming a Word Learner: A Debate on Lexical Acquisition},
  pages 115--135. Oxford University Press, 2000.

\bibitem[Barsalou et~al.(2008)]{barsalou2008grounded}
L.~W. Barsalou et~al.
\newblock Grounded cognition.
\newblock \emph{Annual Review of Psychology}, 59\penalty0 (1):\penalty0
  617--645, 2008.

\bibitem[Baumberger et~al.(2017)Baumberger, Beisbart, and
  Brun]{baumberger2017understanding}
C.~Baumberger, C.~Beisbart, and G.~Brun.
\newblock What is understanding? {A}n overview of recent debates in
  epistemology and philosophy of science.
\newblock In \emph{Explaining Understanding: New Perspectives from Epistemology
  and Philosophy of Science}, pages 1--34. Routledge, 2017.

\bibitem[Bender et~al.(2017)Bender, Beller, and Medin]{bender2017causal}
A.~Bender, S.~Beller, and D.~L. Medin.
\newblock Causal cognition and culture.
\newblock In \emph{The Oxford Handbook of Causal Reasoning}, pages 717--738.
  Oxford University Press, 2017.

\bibitem[Bender and Koller(2020)]{bender2020climbing}
E.~M. Bender and A.~Koller.
\newblock Climbing towards {NLU}: On meaning, form, and understanding in the
  age of data.
\newblock In \emph{Proceedings of the 58th Annual Meeting of the Association
  for Computational Linguistics}, pages 5185--5198, 2020.

\bibitem[Bender et~al.(2021)Bender, Gebru, McMillan-Major, and
  Shmitchell]{bender2021dangers}
E.~M. Bender, T.~Gebru, A.~McMillan-Major, and S.~Shmitchell.
\newblock On the dangers of stochastic parrots: Can language models be too big?
\newblock In \emph{Proceedings of the 2021 ACM Conference on Fairness,
  Accountability, and Transparency}, pages 610--623, 2021.

\bibitem[Binz and Schulz(2022)]{binz2022using}
M.~Binz and E.~Schulz.
\newblock Using cognitive psychology to understand gpt-3, 2022.
\newblock arXiv:2206.14576.

\bibitem[Bommasani et~al.(2021)Bommasani, Hudson, Adeli, Altman, Arora, von
  Arx, Bernstein, Bohg, Bosselut, Brunskill,
  et~al.]{bommasani2021opportunities}
R.~Bommasani, D.~A. Hudson, E.~Adeli, R.~Altman, S.~Arora, S.~von Arx, M.~S.
  Bernstein, J.~Bohg, A.~Bosselut, E.~Brunskill, et~al.
\newblock On the opportunities and risks of foundation models, 2021.
\newblock arXiv:2108.07258.

\bibitem[Brown et~al.(2020)Brown, Mann, Ryder, Subbiah, Kaplan, Dhariwal,
  Neelakantan, Shyam, Sastry, Askell, et~al.]{brown2020language}
T.~Brown, B.~Mann, N.~Ryder, M.~Subbiah, J.~D. Kaplan, P.~Dhariwal,
  A.~Neelakantan, P.~Shyam, G.~Sastry, A.~Askell, et~al.
\newblock Language models are few-shot learners.
\newblock In \emph{Advances in Neural Information Processing Systems},
  volume~33, pages 1877--1901, 2020.

\bibitem[Browning and LeCun(2022)]{browning2022limits}
J.~Browning and Y.~LeCun.
\newblock {AI} and the limits of language, 2022.
\newblock Noema, August 23,
  \url{https://www.noemamag.com/ai-and-the-limits-of-language}.

\bibitem[Carey(1995)]{carey1995origin}
S.~Carey.
\newblock On the origin of causal understanding.
\newblock In D.~Sperber, D.~Premack, and A.~J. Premack, editors, \emph{Causal
  Cognition: A Multidisciplinary Debate}, page 268–308. Clarendon
  Press/Oxford University Press, 1995.

\bibitem[Choudhury et~al.(2022)Choudhury, Rogers, and
  Augenstein]{choudhury2022machine}
S.~R. Choudhury, A.~Rogers, and I.~Augenstein.
\newblock Machine reading, fast and slow: When do models `understand'
  language?, 2022.
\newblock arXiv:2209.07430.

\bibitem[Chowdhery et~al.(2022)Chowdhery, Narang, Devlin, Bosma, Mishra,
  Roberts, Barham, Chung, Sutton, Gehrmann, et~al.]{chowdhery2022palm}
A.~Chowdhery, S.~Narang, J.~Devlin, M.~Bosma, G.~Mishra, A.~Roberts, P.~Barham,
  H.~W. Chung, C.~Sutton, S.~Gehrmann, et~al.
\newblock {PaLM}: Scaling language modeling with {P}athways, 2022.
\newblock arXiv:2204.02311.

\bibitem[Dasgupta et~al.(2022)Dasgupta, Lampinen, Chan, Creswell, Kumaran,
  McClelland, and Hill]{dasgupta2022language}
I.~Dasgupta, A.~K. Lampinen, S.~C.~Y. Chan, A.~Creswell, D.~Kumaran, J.~L.
  McClelland, and F.~Hill.
\newblock Language models show human-like content effects on reasoning, 2022.
\newblock arXiv:2207.07051.

\bibitem[de~Freitas(2022)]{defreitas2022tweet}
N.~de~Freitas, 2022.
\newblock May 14, \url{https://twitter.com/NandoDF/status/1525397036325019649}.

\bibitem[De~Regt(2004)]{de2004discussion}
H.~W. De~Regt.
\newblock Discussion note: Making sense of understanding.
\newblock \emph{Philosophy of Science}, 71\penalty0 (1):\penalty0 98--109,
  2004.

\bibitem[De~Villiers and de~Villiers(2014)]{de2014role}
J.~G. De~Villiers and P.~A. de~Villiers.
\newblock The role of language in theory of mind development.
\newblock \emph{Topics in Language Disorders}, 34\penalty0 (4):\penalty0
  313--328, 2014.

\bibitem[Devlin et~al.(2019)Devlin, Chang, Lee, and Toutanova]{devlin2019bert}
J.~Devlin, M.-W. Chang, K.~Lee, and K.~Toutanova.
\newblock {BERT}: Pre-training of deep bidirectional transformers for language
  understanding.
\newblock In \emph{Proceedings of the 2019 Conference of the North American
  Chapter of the Association for Computational Linguistics: Human Language
  Technologies}, page 4171–4186, 2019.

\bibitem[Dimakis(2022)]{dimakis2022tweet}
A.~Dimakis, 2022.
\newblock May 16,
  \url{https://twitter.com/AlexGDimakis/status/1526388274348150784}.

\bibitem[Dove(2020)]{dove2020more}
G.~Dove.
\newblock More than a scaffold: Language is a neuroenhancement.
\newblock \emph{Cognitive Neuropsychology}, 37\penalty0 (5-6):\penalty0
  288--311, 2020.

\bibitem[Gardner et~al.(2021)Gardner, Merrill, Dodge, Peters, Ross, Singh, and
  Smith]{gardner2021competency}
M.~Gardner, W.~Merrill, J.~Dodge, M.~E. Peters, A.~Ross, S.~Singh, and
  N.~Smith.
\newblock Competency problems: On finding and removing artifacts in language
  data.
\newblock In \emph{Proceedings of the 2021 Conference on Empirical Methods in
  Natural Language Processing}, 2021.

\bibitem[Geirhos et~al.(2020)Geirhos, Jacobsen, Michaelis, Zemel, Brendel,
  Bethge, and Wichmann]{geirhos2020shortcut}
R.~Geirhos, J.-H. Jacobsen, C.~Michaelis, R.~Zemel, W.~Brendel, M.~Bethge, and
  F.~A. Wichmann.
\newblock Shortcut learning in deep neural networks.
\newblock \emph{Nature Machine Intelligence}, 2\penalty0 (11):\penalty0
  665--673, 2020.

\bibitem[Gelman(2009)]{gelman2009learning}
S.~A. Gelman.
\newblock Learning from others: Children's construction of concepts.
\newblock \emph{Annual Review of Psychology}, 60:\penalty0 115--140, 2009.

\bibitem[George et~al.(2020)George, L{\'a}zaro-Gredilla, and
  Guntupalli]{george2020captcha}
D.~George, M.~L{\'a}zaro-Gredilla, and J.~S. Guntupalli.
\newblock From {CAPTCHA} to commonsense: How brain can teach us about
  artificial intelligence.
\newblock \emph{Frontiers in Computational Neuroscience}, 14:\penalty0 554097,
  2020.

\bibitem[Gibbs(2017)]{gibbs2017metaphor}
R.~W. Gibbs.
\newblock \emph{Metaphor Wars}.
\newblock Cambridge University Press, 2017.

\bibitem[Goldwater and Gentner(2015)]{goldwater2015acquisition}
M.~B. Goldwater and D.~Gentner.
\newblock On the acquisition of abstract knowledge: Structural alignment and
  explication in learning causal system categories.
\newblock \emph{Cognition}, 137:\penalty0 137--153, 2015.

\bibitem[Goodman et~al.(2011)Goodman, Ullman, and
  Tenenbaum]{goodman2011learning}
N.~D. Goodman, T.~D. Ullman, and J.~B. Tenenbaum.
\newblock Learning a theory of causality.
\newblock \emph{Psychological Review}, 118\penalty0 (1):\penalty0 110, 2011.

\bibitem[Gopnik(2011)]{gopnik2011unified}
A.~Gopnik.
\newblock A unified account of abstract structure and conceptual change:
  Probabilistic models and early learning mechanisms.
\newblock \emph{Behavioral and Brain Sciences}, 34\penalty0 (3):\penalty0 129,
  2011.

\bibitem[Gopnik(2022{\natexlab{a}})]{gopnik2022causal}
A.~Gopnik.
\newblock Causal models and cognitive development.
\newblock In H.~Geffner, R.~Dechter, and J.~Y. Halpern, editors,
  \emph{Probabilistic and Causal Inference: The Works of Judea Pearl}, pages
  593--604. Association for Computing Machinery, 2022{\natexlab{a}}.

\bibitem[Gopnik(2022{\natexlab{b}})]{gopnikAI2022}
A.~Gopnik.
\newblock What {AI} still doesn’t know how to do, 2022{\natexlab{b}}.
\newblock Wall Street Journal, July 15,
  \url{https://www.wsj.com/articles/what-ai-still-doesnt-know-how-to-do-11657891316}.

\bibitem[Gopnik and Wellman(1994)]{gopnik1994theory}
A.~Gopnik and H.~M. Wellman.
\newblock The theory theory.
\newblock In \emph{Domain Specificity in Cognition and Culture}, pages
  257--293. 1994.

\bibitem[Gururangan et~al.(2018)Gururangan, Swayamdipta, Levy, Schwartz,
  Bowman, and Smith]{gururangan2018annotation}
S.~Gururangan, S.~Swayamdipta, O.~Levy, R.~Schwartz, S.~R. Bowman, and N.~A.
  Smith.
\newblock Annotation artifacts in natural language inference data.
\newblock In \emph{Proceedings of the 2018 Conference of the North American
  Chapter of the Association for Computational Linguistics: Human Language
  Technologies}, pages 107--112, 2018.

\bibitem[Habernal et~al.(2018)Habernal, Wachsmuth, Gurevych, and
  Stein]{habernal2018argument}
I.~Habernal, H.~Wachsmuth, I.~Gurevych, and B.~Stein.
\newblock The argument reasoning comprehension task: Identification and
  reconstruction of implicit warrants.
\newblock In \emph{Proceedings of the 2018 Conference of the North American
  Chapter of the Association for Computational Linguistics: Human Language
  Technologies}, page 1930–1940, 2018.

\bibitem[Hofstadter(1995)]{hofstadter1995ineradicable}
D.~R. Hofstadter.
\newblock \emph{Fluid Concepts and Creative Analogies: Computer Models of the
  Fundamental Mechanisms of Thought}.
\newblock Basic Books, 1995.
\newblock Preface to Chapter 4.

\bibitem[Hofstadter and Sander(2013)]{hofstadter2013surfaces}
D.~R. Hofstadter and E.~Sander.
\newblock \emph{Surfaces and Essences: Analogy as the Fuel and Fire of
  Thinking}.
\newblock Basic books, 2013.

\bibitem[Jones and Thornton(2022)]{jones2022impact}
D.~T. Jones and J.~M. Thornton.
\newblock The impact of {AlphaFold2} one year on.
\newblock \emph{Nature Methods}, 19\penalty0 (1):\penalty0 15--20, 2022.

\bibitem[Jumper et~al.(2021)Jumper, Evans, Pritzel, Green, Figurnov,
  Ronneberger, Tunyasuvunakool, Bates, {\v{Z}}{\'\i}dek, Potapenko,
  et~al.]{jumper2021highly}
J.~Jumper, R.~Evans, A.~Pritzel, T.~Green, M.~Figurnov, O.~Ronneberger,
  K.~Tunyasuvunakool, R.~Bates, A.~{\v{Z}}{\'\i}dek, A.~Potapenko, et~al.
\newblock Highly accurate protein structure prediction with {AlphaFold}.
\newblock \emph{Nature}, 596\penalty0 (7873):\penalty0 583--589, 2021.

\bibitem[Keil(2006)]{keil2006explanation}
F.~C. Keil.
\newblock Explanation and understanding.
\newblock \emph{Annual Review of Psychology}, 57:\penalty0 227, 2006.

\bibitem[Krakauer(2020)]{krakauer2020limits}
D.~C. Krakauer.
\newblock At the limits of thought, 2020.
\newblock Aeon, April 20,
  \url{https://aeon.co/essays/will-brains-or-algorithms-rule-the-kingdom-of-science}.

\bibitem[Kvanvig(2018)]{kvanvig2018knowledge}
J.~L. Kvanvig.
\newblock Knowledge, understanding, and reasons for belief.
\newblock In \emph{{The Oxford Handbook of Reasons and Normativity}}, page
  685–705. Oxford University Press, 2018.

\bibitem[Lake and Murphy(2021)]{lake2021word}
B.~M. Lake and G.~L. Murphy.
\newblock Word meaning in minds and machines.
\newblock \emph{Psychological Review}, 2021.

\bibitem[Lake et~al.(2017)Lake, Ullman, Tenenbaum, and
  Gershman]{lake2017building}
B.~M. Lake, T.~D. Ullman, J.~B. Tenenbaum, and S.~J. Gershman.
\newblock Building machines that learn and think like people.
\newblock \emph{Behavioral and Brain Sciences}, 40, 2017.

\bibitem[Lakoff and Johnson(1980)]{lakoff1980metaphorical}
G.~Lakoff and M.~Johnson.
\newblock The metaphorical structure of the human conceptual system.
\newblock \emph{Cognitive Science}, 4\penalty0 (2):\penalty0 195--208, 1980.

\bibitem[Lapuschkin et~al.(2019)Lapuschkin, W{\"a}ldchen, Binder, Montavon,
  Samek, and M{\"u}ller]{lapuschkin2019unmasking}
S.~Lapuschkin, S.~W{\"a}ldchen, A.~Binder, G.~Montavon, W.~Samek, and K.-R.
  M{\"u}ller.
\newblock Unmasking {Clever Hans} predictors and assessing what machines really
  learn.
\newblock \emph{Nature Communications}, 10\penalty0 (1):\penalty0 1--8, 2019.

\bibitem[Laverghetta et~al.(2022)Laverghetta, Nighojkar, Mirzakhalov, and
  Licato]{laverghetta2022predicting}
A.~Laverghetta, A.~Nighojkar, J.~Mirzakhalov, and J.~Licato.
\newblock Predicting human psychometric properties using computational language
  models.
\newblock In \emph{Annual Meeting of the Psychometric Society}, pages 151--169.
  Springer, 2022.

\bibitem[Li et~al.(2021)Li, Nye, and Andreas]{li2021implicit}
B.~Z. Li, M.~Nye, and J.~Andreas.
\newblock Implicit representations of meaning in neural language models.
\newblock In \emph{Proceedings of the 59th Annual Meeting of the Association
  for Computational Linguistics}, page 1813–1827, 2021.

\bibitem[Linzen(2020)]{linzen2020can}
T.~Linzen.
\newblock How can we accelerate progress towards human-like linguistic
  generalization?
\newblock In \emph{In Proceedings of the 58th Annual Meeting of the Association
  for Computational Linguistics}, page 5210–17, 2020.

\bibitem[Lupyan and Bergen(2016)]{lupyan2016language}
G.~Lupyan and B.~Bergen.
\newblock How language programs the mind.
\newblock \emph{Topics in Cognitive Science}, 8\penalty0 (2):\penalty0
  408--424, 2016.

\bibitem[Mahowald et~al.(2023)Mahowald, Ivanova, Blank, Kanwisher, Tenenbaum,
  and Fedorenko]{mahowald2023dissociating}
K.~Mahowald, A.~A. Ivanova, I.~A. Blank, N.~Kanwisher, J.~B. Tenenbaum, and
  E.~Fedorenko.
\newblock Dissociating language and thought in large language models: a
  cognitive perspective, 2023.
\newblock arXiv:2301.06627.

\bibitem[Mandler(1992)]{mandler1992build}
J.~M. Mandler.
\newblock How to build a baby: {II. C}onceptual primitives.
\newblock \emph{Psychological Review}, 99\penalty0 (4):\penalty0 587, 1992.

\bibitem[Manning(2022)]{manning2022understanding}
C.~D. Manning.
\newblock Human language understanding and reasoning.
\newblock \emph{Daedalus}, 151\penalty0 (2):\penalty0 127--138, 2022.

\bibitem[Marcus(2022)]{marcus2022nonsense}
G.~Marcus.
\newblock Nonsense on stilts, 2022.
\newblock Substack, June 12,
  \url{https://garymarcus.substack.com/p/nonsense-on-stilts}.

\bibitem[McCoy et~al.(2019)McCoy, Pavlick, and Linzen]{mccoy2019right}
R.~T. McCoy, E.~Pavlick, and T.~Linzen.
\newblock Right for the wrong reasons: Diagnosing syntactic heuristics in
  natural language inference.
\newblock In \emph{Proceedings of the 57th Annual Meeting of the Association
  for Computational Linguistics}, page 3428–3448, 2019.

\bibitem[Michael et~al.(2022)Michael, Holtzman, Parrish, Mueller, Wang, Chen,
  Madaan, Nangia, Pang, Phang, et~al.]{michael2022nlp}
J.~Michael, A.~Holtzman, A.~Parrish, A.~Mueller, A.~Wang, A.~Chen, D.~Madaan,
  N.~Nangia, R.~Y. Pang, J.~Phang, et~al.
\newblock What do {NLP} researchers believe? {R}esults of the {NLP} community
  metasurvey, 2022.
\newblock arXiv:2208.12852.

\bibitem[Min et~al.(2021)Min, Ross, Sulem, Veyseh, Nguyen, Sainz, Agirre,
  Heinz, and Roth]{min2021recent}
B.~Min, H.~Ross, E.~Sulem, A.~P.~B. Veyseh, T.~H. Nguyen, O.~Sainz, E.~Agirre,
  I.~Heinz, and D.~Roth.
\newblock Recent advances in natural language processing via large pre-trained
  language models: A survey, 2021.
\newblock arXiv:2111.01243.

\bibitem[Mitchell(2019)]{mitchell2019artificial}
M.~Mitchell.
\newblock Artificial intelligence hits the barrier of meaning.
\newblock \emph{Information}, 10\penalty0 (2):\penalty0 51, 2019.

\bibitem[Morris et~al.(2003)Morris, Menon, and Ames]{morris2003culturally}
M.~W. Morris, T.~Menon, and D.~R. Ames.
\newblock Culturally conferred conceptions of agency: A key to social
  perception of persons, groups, and other actors.
\newblock In \emph{Personality and Social Psychology Review}, pages 169--182.
  Psychology Press, 2003.

\bibitem[Murphy(1996)]{murphy1996metaphoric}
G.~L. Murphy.
\newblock On metaphoric representation.
\newblock \emph{Cognition}, 60\penalty0 (2):\penalty0 173--204, 1996.

\bibitem[Niven and Kao(2019)]{niven2019probing}
T.~Niven and H.-Y. Kao.
\newblock Probing neural network comprehension of natural language arguments.
\newblock In \emph{Proceedings of the 57th Annual Meeting of the Association
  for Computational Linguistics}, pages 4658--4664, 2019.

\bibitem[Norenzayan and Nisbett(2000)]{norenzayan2000culture}
A.~Norenzayan and R.~E. Nisbett.
\newblock Culture and causal cognition.
\newblock \emph{Current Directions in Psychological Science}, 9\penalty0
  (4):\penalty0 132--135, 2000.

\bibitem[Olsson et~al.(2022)Olsson, Elhage, Nanda, Joseph, DasSarma, Henighan,
  Mann, Askell, Bai, Chen, et~al.]{olsson2022context}
C.~Olsson, N.~Elhage, N.~Nanda, N.~Joseph, N.~DasSarma, T.~Henighan, B.~Mann,
  A.~Askell, Y.~Bai, A.~Chen, et~al.
\newblock In-context learning and induction heads, 2022.
\newblock arXiv preprint arXiv:2209.11895.

\bibitem[Ouyang et~al.(2022)Ouyang, Wu, Jiang, Almeida, Wainwright, Mishkin,
  Zhang, Agarwal, Slama, Ray, et~al.]{ouyang2022training}
L.~Ouyang, J.~Wu, X.~Jiang, D.~Almeida, C.~L. Wainwright, P.~Mishkin, C.~Zhang,
  S.~Agarwal, K.~Slama, A.~Ray, et~al.
\newblock Training language models to follow instructions with human feedback,
  2022.
\newblock arXiv:2203.02155.

\bibitem[Pearl(2018)]{pearl2018theoretical}
J.~Pearl.
\newblock Theoretical impediments to machine learning with seven sparks from
  the causal revolution, 2018.
\newblock arXiv:1801.04016.

\bibitem[Piantasodi and Hill(2022)]{piantasodi2022meaning}
S.~T. Piantasodi and F.~Hill.
\newblock Meaning without reference in large language models, 2022.
\newblock arXiv:2208.02957.

\bibitem[Sadler and Regan(2019)]{sadler2019game}
M.~Sadler and N.~Regan.
\newblock \emph{Game changer: {AlphaZero}’s Groundbreaking Chess Strategies
  and the Promise of {AI}}.
\newblock Alkmaar, 2019.

\bibitem[Schulman et~al.(2022)Schulman, Zoph, Kim, Hilton, Menick, Weng, Uribe,
  Fedus, Metz, Pokorny, et~al.]{schulman2022chatgpt}
J.~Schulman, B.~Zoph, C.~Kim, J.~Hilton, J.~Menick, J.~Weng, J.~Uribe,
  L.~Fedus, L.~Metz, M.~Pokorny, et~al.
\newblock Chat{GPT}: Optimizing language models for dialogue, 2022.
\newblock November 30, \url{https://openai.com/blog/chatgpt}.

\bibitem[Sejnowski(2022)]{sejnowski2022large}
T.~Sejnowski.
\newblock Large language models and the reverse {T}uring test, 2022.
\newblock arXiv:2207.14382.

\bibitem[Shanahan(2022)]{shanahan2022talking}
M.~Shanahan.
\newblock Talking about large language models, 2022.
\newblock arXiv:2212.03551.

\bibitem[Silver et~al.(2017)Silver, Hubert, Schrittwieser, Antonoglou, Lai,
  Guez, Lanctot, Sifre, Kumaran, Graepel, et~al.]{silver2017mastering}
D.~Silver, T.~Hubert, J.~Schrittwieser, I.~Antonoglou, M.~Lai, A.~Guez,
  M.~Lanctot, L.~Sifre, D.~Kumaran, T.~Graepel, et~al.
\newblock Mastering chess and shogi by self-play with a general reinforcement
  learning algorithm, 2017.
\newblock arXiv:1712.01815.

\bibitem[Sloman and Lagnado(2015)]{sloman2015causality}
S.~A. Sloman and D.~Lagnado.
\newblock Causality in thought.
\newblock \emph{Annual Review of Psychology}, 66:\penalty0 223--247, 2015.

\bibitem[Smolensky et~al.(2022)Smolensky, McCoy, Fernandez, Goldrick, and
  Gao]{smolensky2022neurocompositional}
P.~Smolensky, R.~McCoy, R.~Fernandez, M.~Goldrick, and J.~Gao.
\newblock Neurocompositional computing: From the central paradox of cognition
  to a new generation of {AI} systems.
\newblock \emph{AI Magazine}, 43\penalty0 (3):\penalty0 308--322, 2022.

\bibitem[Spelke and Kinzler(2007)]{spelke2007core}
E.~S. Spelke and K.~D. Kinzler.
\newblock Core knowledge.
\newblock \emph{Developmental Science}, 10\penalty0 (1):\penalty0 89--96, 2007.

\bibitem[Strevens(2013)]{strevens2013no}
M.~Strevens.
\newblock No understanding without explanation.
\newblock \emph{Studies in History and Philosophy of Science Part A},
  44\penalty0 (3):\penalty0 510--515, 2013.

\bibitem[Thoppilan et~al.(2022)Thoppilan, De~Freitas, Hall, Shazeer,
  Kulshreshtha, Cheng, Jin, Bos, Baker, Du, et~al.]{thoppilan2022lamda}
R.~Thoppilan, D.~De~Freitas, J.~Hall, N.~Shazeer, A.~Kulshreshtha, H.-T. Cheng,
  A.~Jin, T.~Bos, L.~Baker, Y.~Du, et~al.
\newblock {LaMDA}: Language models for dialog applications, 2022.
\newblock arXiv:2201.08239.

\bibitem[Trott et~al.(2022)Trott, Jones, Chang, Michaelov, and
  Bergen]{trott2022large}
S.~Trott, C.~Jones, T.~Chang, J.~Michaelov, and B.~Bergen.
\newblock Do large language models know what humans know?, 2022.
\newblock arXiv:2209.01515.

\bibitem[Wang et~al.(2018)Wang, Singh, Michael, Hill, Levy, and
  Bowman]{wang2018glue}
A.~Wang, A.~Singh, J.~Michael, F.~Hill, O.~Levy, and S.~R. Bowman.
\newblock {GLUE}: A multi-task benchmark and analysis platform for natural
  language understanding.
\newblock In \emph{Proceedings of the 2018 EMNLP Workshop BlackboxNLP:
  Analyzing and Interpreting Neural Networks for NLP}, pages 353--355.
  Association for Computational Linguistics, 2018.

\bibitem[Wang et~al.(2019)Wang, Pruksachatkun, Nangia, Singh, Michael, Hill,
  Levy, and Bowman]{wang2019superglue}
A.~Wang, Y.~Pruksachatkun, N.~Nangia, A.~Singh, J.~Michael, F.~Hill, O.~Levy,
  and S.~R. Bowman.
\newblock {{SuperGLUE}: A stickier benchmark for general-purpose language
  understanding systems}.
\newblock In \emph{Advances in Neural Information Processing Systems},
  volume~32, pages 3266--3280, 2019.

\bibitem[Waxman and Gelman(2009)]{waxman2009early}
S.~R. Waxman and S.~A. Gelman.
\newblock Early word-learning entails reference, not merely associations.
\newblock \emph{Trends in Cognitive Sciences}, 13\penalty0 (6):\penalty0
  258--263, 2009.

\bibitem[Wei et~al.(2022{\natexlab{a}})Wei, Tay, Bommasani, Raffel, Zoph,
  Borgeaud, Yogatama, Bosma, Zhou, Metzler, et~al.]{wei2022emergent}
J.~Wei, Y.~Tay, R.~Bommasani, C.~Raffel, B.~Zoph, S.~Borgeaud, D.~Yogatama,
  M.~Bosma, D.~Zhou, D.~Metzler, et~al.
\newblock Emergent abilities of large language models, 2022{\natexlab{a}}.
\newblock arXiv:2206.07682.

\bibitem[Wei et~al.(2022{\natexlab{b}})Wei, Wang, Schuurmans, Bosma, Chi, Le,
  and Zhou]{wei2022chain}
J.~Wei, X.~Wang, D.~Schuurmans, M.~Bosma, E.~Chi, Q.~Le, and D.~Zhou.
\newblock Chain of thought prompting elicits reasoning in large language
  models, 2022{\natexlab{b}}.
\newblock arXiv:2201.11903.

\bibitem[Weizenbaum(1976)]{weizenbaum1976computer}
J.~Weizenbaum.
\newblock \emph{Computer Power and Human Reason: From Judgment to Calculation.}
\newblock WH Freeman \& Co, 1976.

\bibitem[Wellman and Gelman(1992)]{wellman1992cognitive}
H.~M. Wellman and S.~A. Gelman.
\newblock Cognitive development: Foundational theories of core domains.
\newblock \emph{Annual Review of Psychology}, 43\penalty0 (1):\penalty0
  337--375, 1992.

\end{thebibliography}

\end{document}